\documentclass[10pt,twocolumn,letterpaper]{article}

\usepackage{iccv}
\usepackage{times}
\usepackage{epsfig}
\usepackage{graphicx}
\usepackage{amsmath}
\usepackage{amssymb}
\usepackage{caption}
\usepackage{color}
\usepackage{makecell}
\usepackage[titletoc]{appendix}
\usepackage{xspace}
\usepackage{color}
\usepackage{multirow}
\usepackage{graphics}
\usepackage{adjustbox}
\usepackage{subfigure}
\usepackage{amsmath}
\usepackage{paralist} 
\usepackage{enumitem} 
\usepackage{booktabs} 
\usepackage{array}

\graphicspath{{figures/}}

\long\def\ignorethis#1{}

\usepackage{tabularx}

\definecolor{gray}{rgb}{0.5,0.5,0.5}
\definecolor{MyBlue}{rgb}{0,0,1.0}
\definecolor{MyYellow}{rgb}{0.9,0.9,0}
\definecolor{MyRed}{rgb}{0.8,0.2,0}
\definecolor{MyGreen}{rgb}{0,0.5,0.0}
\definecolor{MyGray}{rgb}{0.4,0.4,0.4}

\newlength\paramargin
\newlength\figmargin
\newlength\secmargin

\newcolumntype{L}[1]{>{\raggedright\let\newline\\\arraybackslash\hspace{0pt}}m{#1}}
\newcolumntype{C}[1]{>{\centering\let\newline\\\arraybackslash\hspace{0pt}}m{#1}}
\newcolumntype{R}[1]{>{\raggedleft\let\newline\\\arraybackslash\hspace{0pt}}m{#1}}

\def\ie{i.e.,~}
\def\eg{e.g.,~}

\usepackage{url}

\def\ie{i.e.,~}
\def\eg{e.g.,~}

\usepackage[pagebackref=true,breaklinks=true,letterpaper=true,colorlinks,bookmarks=false]{hyperref}

\iccvfinalcopy 

\def\iccvPaperID{****} 

\ificcvfinal\fi

\title{CFNet: Learning Correlation Functions for One-Stage Panoptic Segmentation}

\author{
Yifeng Chen$^{1}$\thanks{Equal Contribution.} ,
Wenqing Chu$^{2}$\footnotemark[1] ,
Fangfang Wang$^{1}$, Ying Tai$^{2}$, \\
Ran Yi$^{3}$,
Zhenye Gan$^{2}$,
Liang Yao$^{2}$,
Chengjie Wang$^{2}$,
Xi Li$^{1}$\\
$^{1}$Zhejiang University,
$^{2}$Tencent Youtu Lab,
$^{3}$Shanghai Jiao Tong University
}

\begin{document}
\maketitle

\begin{abstract}

Recently, there is growing attention on one-stage panoptic segmentation methods which aim to segment instances and stuff jointly within a fully convolutional pipeline efficiently.
However, most of the existing works directly feed the backbone features to various segmentation heads ignoring the demands for semantic and instance segmentation are different: The former needs semantic-level discriminative features, while
the latter requires features to be distinguishable across instances.
To alleviate this, we propose to first predict semantic-level and instance-level correlations among different locations that are utilized to enhance the backbone features, and then feed the improved discriminative features into the corresponding segmentation heads, respectively.
Specifically, we organize the correlations between
a given location and all locations as a continuous sequence and predict it as a whole.
Considering that such a sequence can be
extremely complicated, we adopt Discrete Fourier Transform (DFT), a tool that can approximate
an arbitrary sequence parameterized by amplitudes and phrases.
For different tasks, we generate these parameters from the backbone features in a fully convolutional way which is optimized implicitly by corresponding tasks. 
As a result, these accurate and consistent correlations contribute to producing plausible discriminative features which meet the requirements of the complicated panoptic segmentation task.
To verify the effectiveness of our methods, we conduct experiments on several challenging panoptic segmentation datasets and achieve state-of-the-art performance on MS COCO with $45.1$\% PQ and ADE20k with $32.6$\% PQ.
\end{abstract}
\vspace{-2mm}
\section{Introduction}
Panoptic segmentation~\cite{kirillov2019panoptic1}, a fundamental and challenging problem in scene parsing, is to carry out instance segmentation for foreground things and semantic segmentation for background stuff.
In recent years, a large number of approaches~\cite{kirillov2019panoptic2, sofiiuk2019adaptIS,zhang2021ada,du2021save,ren2021refine} have been proposed for panoptic segmentation due to its wide potential applications in autonomous driving, robotics, and image editing.
Among these algorithms, the workflow of one-stage methods~\cite{chen2019pdeeplab,li2021panopticfcn,hong2021lpsnet} also referred to box-free is simple and effective which attracts large attention.

\begin{figure}[t]
	\centering
	\includegraphics[width=1.0\linewidth]{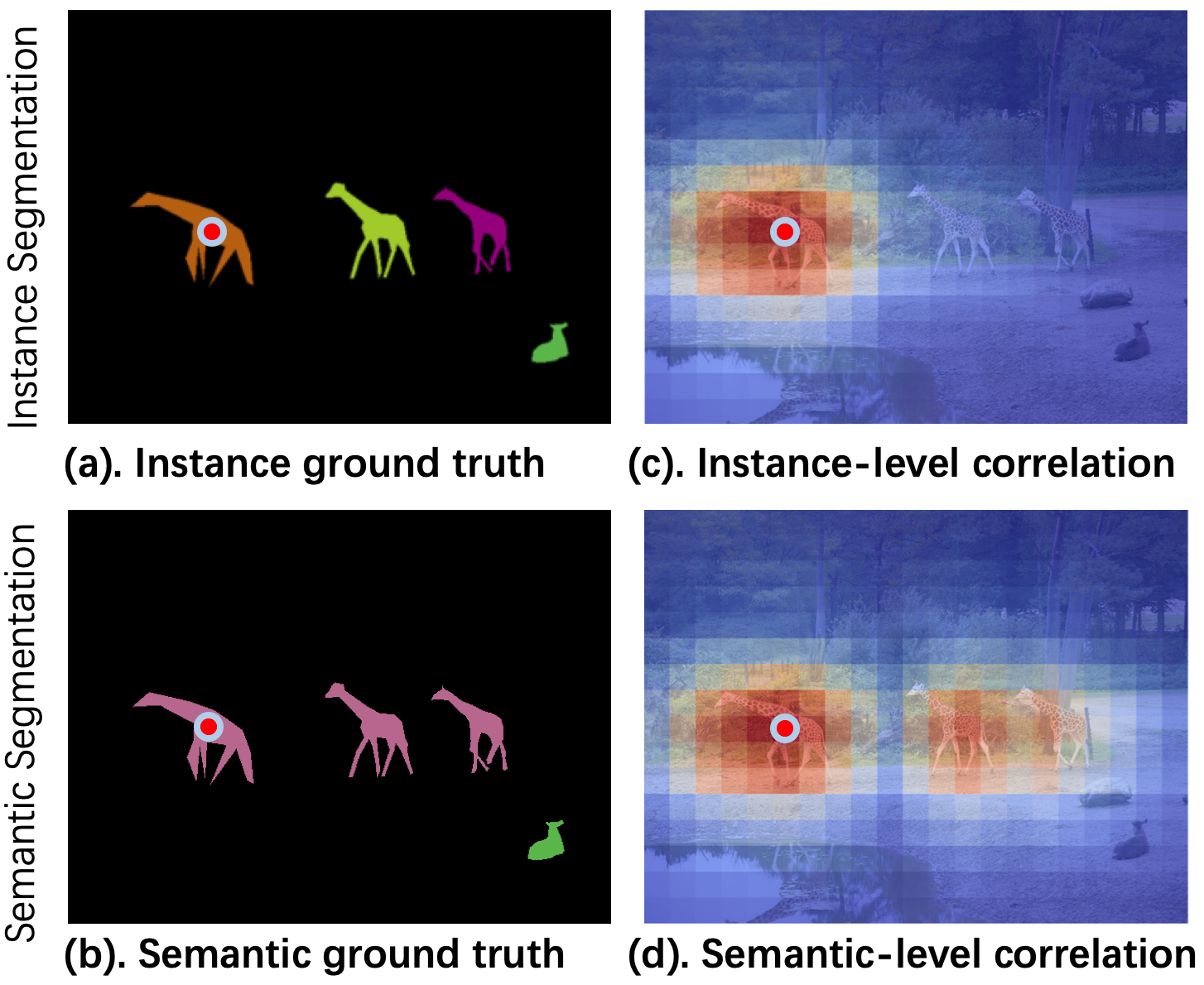}
	\caption{
		Visualizations of different correlations with respect to a given point~(red point), where red color represents higher correlation. (a), (b) are ground truth for instance and semantic segmentation. (c), (d) are correlations modeled by the proposed function. Our mechanism is able to model correlations from scenes directly, showing a scene-related task-specific result in (c) and (d).
	}
	\label{fig:intro}
	\vspace{-2.0em}
\end{figure}

Current one-stage methods usually leverage a Feature Pyramid Network~\cite{lin2017feature} based backbone to extract feature maps and then attach different segmentation heads to them.
However, we note that the demands for semantic and instance segmentation are different: The former needs discriminative features across different categories, while the latter requires features to be instance-level distinguishable.
To alleviate this, the attention mechanism is introduced to help semantic segmentation~\cite{wang2018location,carion2020end}, which can enhance the semantic consistency within the same class.
For the instance segmentation, coordinates~\cite{wang2020solo,carion2020end,liu2018intriguing} are typically concatenated with the backbone feature maps for capturing spatial cues which have been proven to be effective for enhancing instance-level discrimination to some degree.
From a unified view, the attention mechanism and CoordConv~\cite{liu2018intriguing} both aim to introduce discriminative features, while the former is for semantic-level and the latter is instance-level. 
Unfortunately, these features are heuristic or fixed that are not flexible insufficient for various scenes, \eg discrimination should be stricter for more crowded scenes in instance segmentation.

In this work, we propose to predict semantic-level and instance-level correlations and then utilize this information to guide the feature enhancement for efficient panoptic segmentation.
Specifically, we treat these correlations as task-driven and scene-relevant so that we could learn them through implicit supervisions from tasks.
As shown in Fig.~\ref{fig:intro}, we organize the correlations between a given location and all locations as a sequence and learn it as a whole.
In other words, we seek to learn a function with coordinates as input to approximate this sequence. 
It is more sensible than learning correlations individually since the latter ignores the fact that two adjacent locations tend to have continuous correlations.
Meanwhile, we note that such a sequence can be extremely complicated, which requires a powerful function family to cover it.
To meet these requirements, we adopt DFT~(Discrete Fourier Transform~\cite{oppenheim1999discrete}), a tool that can approximate an arbitrary sequence to the addition of several sinusoidal base functions parameterized by amplitudes and phrases.
Moreover, we introduce a convolutional module to predict the function parameters for capturing different kinds of correlations in various segmentation tasks. This module leverages the scene content as input and is optimized implicitly by corresponding tasks.

With the predicted correlations, we can improve the backbone features effectively. 
For semantic segmentation, we directly aggregate those pixel features within the same semantic to enhance the discrimination. 
For instance segmentation, we stack the pixel correlations with other locations as discriminative features and concatenate them with the backbone features to improve the discrimination. 
As a result, different segmentation heads could obtain plausible inputs and generate the segmentation results accurately.
%
 
The main contributions of our work are as follows:
\vspace{-2mm}
\begin{itemize}
	\setlength{\itemsep}{0pt}
	\setlength{\parsep}{0pt}
	\setlength{\parskip}{0.5pt}
	\item We introduce a new DFT based correlation function which is able to model task-specific, scene-related correlations among different locations flexibly.
	\item We point out that one-stage panoptic segmentation needs different kinds of feature discrimination and propose to learn the correlation function parameters in a convoluitonal manner which is inserted into the one-stage panoptic segmentation pipeline seamlessly.
	\item We conduct extensive experiments to analyze the proposed method which obtains SOTA panoptic segmentation performance with $45.1$\% PQ on MS COCO and $32.6$\% on ADE20k.
\end{itemize}
\section{Related Work}

\paragraph{Semantic segmentation.}
Semantic segmentation, a task of pixel-wise classification, has seen great progress since the emergence of fully convolutional network~\cite{long2015fcn}.
Various methods have been proposed to make use of beneficial context information.
PSPNet~\cite{zhao2017pyramid} adopts pyramid modules to exploit multi-scale features.
Atrous~\cite{chen2018deeplab, chen2017rethinking, chen2018encoder} and deformable convolutions~\cite{dai2017deformable} improve the sampling process of vanilla convolutions.
The attention mechanism~\cite{vaswani2017attention,huang2019ccnet} considers the pixel-to-pixel relationship as non-local and designs a global feature broadcast scheme based on semantic affinities.
Meanwhile, several methods~\cite{zhao2019correlation,wu2019improving,zhao2019region} introduce explicit correlation based constraints on the semantic predictions to capture image structure information. 

\vspace{-10pt}
\paragraph{Instance segmentation.}
Instance segmentation targets to assign a category and an instance ID to each object pixel.
Two-stage methods, represented by Mask-RCNN~\cite{he2017mask}, generate region proposals in the first stage and segment them in the second stage.
One-stage methods seek ways to model masks densely.
TensorMask~\cite{chen2019tensormask} directly learns instance masks as flattened vectors, while SOLO~\cite{wang2020solo} reduces its complexity by dividing the input into uniform grids.
Contour-based methods~\cite{xie2020polarmask, xu2019ese} parameterize masks by polynomials fitting and learn the internal parameters instead.
Embedding-based methods~\cite{newell2017associative, ying2019embedmask} extract masks by learning pixel and proposal embedding, while CondInst~\cite{tian2020conditional} and SOLOv2~\cite{wang2020solov2} take a further step by introducing dynamic convolutions~\cite{yang2019condconv}.

\vspace{-10pt}
\paragraph{Panoptic segmentation.}
Panoptic segmentation~\cite{kirillov2019panoptic1} is a joint task of instance segmentation and semantic segmentation.
Two-stage methods~\cite{kirillov2019panoptic2, porzi2019seamless} extend Mask-RCNN with a fully convolutional branch and apply a heuristic fusion algorithm~\cite{kirillov2019panoptic1} to fuse instances and stuff predictions.
Lots of efforts have been paid to building connections among tasks to leverage reciprocal information~\cite{chen2020banet, li2019attention, wu2020bidirectional}.
To improve the fusion process, both learnable~\cite{lazarow2020learning, liu2019end, yang2019sognet} and parameter-free~\cite{li2020unifying, xiong2019upsnet} modules are proposed to deal with overlapping.
One-stage methods~\cite{chang2020epsnet,chen2020panonet}, on the other hand, focus on finding efficient ways to segment instances and stuff jointly within a fully convolutional pipeline.
Embedding-based methods~\cite{gao2020learning, hou2020real, liu2020casnet} extract instance masks by comparing the affinity between mask embeddings and instance embeddings.
Voting-based methods~\cite{chen2019pdeeplab, wang2020pixel,wang2020axial} obtain masks through hough voting~\cite{ballard1981generalizing} by learning object centers and pixel-wise offsets.
\cite{wang2019associatively} incorporates semantic features into instance segmentation and leverages instance embeddings for enhancing semantic segmentation.
However, all these instance-level discrimination enhancement modules are static, which will hurt their generalizability to various scenes. 
Instead, we seek to model different correlations adaptively and continuously. 
We propose correlation functions to model such correlation as scene-relevant and improve feature representations in various segmentation tasks.

%
%
\vspace{-2mm}

\section{Proposed Method}
In this section, we describe the proposed efficient one-stage pipeline for panoptic segmentation in detail.
We first show the overall framework and then introduce different components.
To alleviate the conflicts between different segmentation needs, we elaborate on the adapted and continuous correlation functions which are utilized to guide the learning for task-specific discriminative features.
After that, two applications of correlations, namely Semantic Correlation Module~(SCM) and Instance Correlation Module~(ICM), are designed respectively for semantic and instance segmentation.

\subsection{Architecture}
\label{sec:3.1}
\begin{figure}[tb]
	\centering
	\includegraphics[width=1.0\linewidth]{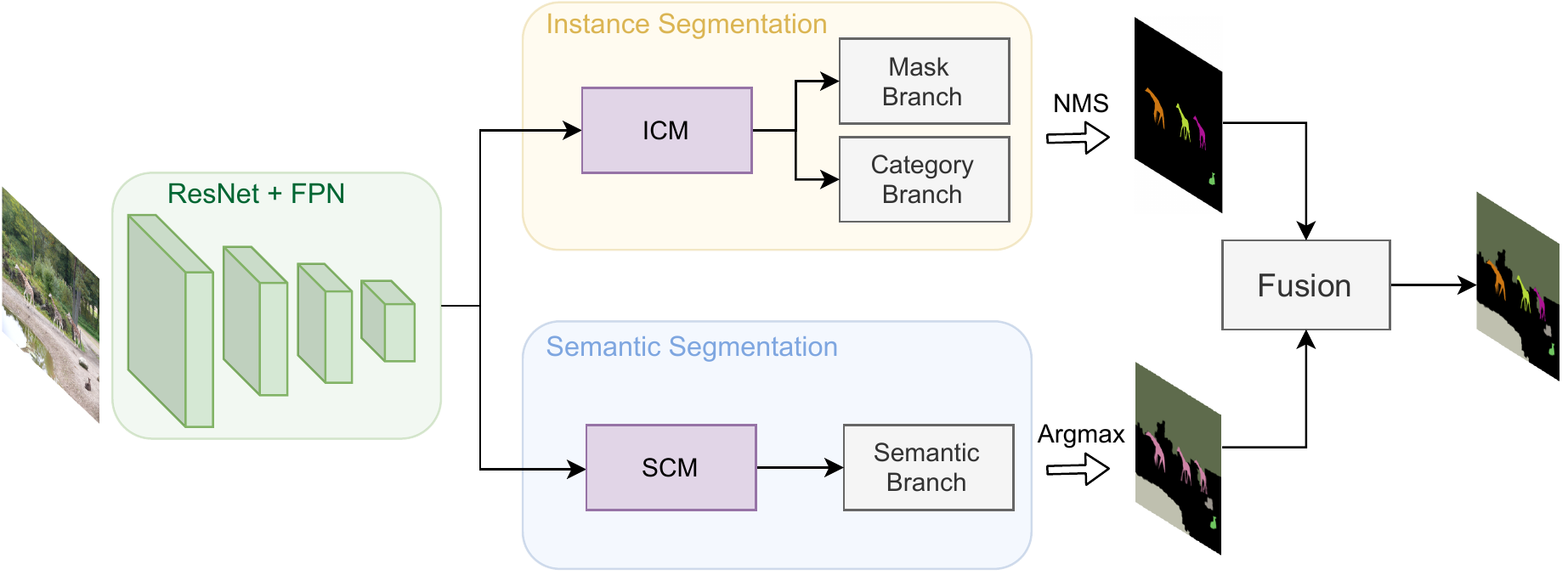}
	\caption{
		Architecture of the proposed one-stage panoptic segmentation pipeline, which contains four parts: a backbone feature extractor, an instance segmentation branch with ICM, a semantic segmentation branch with SCM, and a post-processing step for generating segmentation results.	
	}
	\label{fig:architecture}
	\vspace{-1.0em}
\end{figure}

To pursue both speed and accuracy, we employ a single-stage pipeline for panoptic segmentation, which contains four parts, including a backbone feature extractor, an instance segmentation branch with ICM and a semantic segmentation branch with SCM, and a post-processing algorithm to obtain panoptic results.
The overall architecture of our method is shown in Fig.~\ref{fig:architecture}.

\vspace{-10pt}
\paragraph{Backbone feature extractor.}
Backbone feature extractor aims to extract shared features for subsequent tasks.
We adopt ResNet~\cite{he2016deep}+FPN~\cite{lin2017feature} as the backbone network, where the outputs from different blocks of a ResNet are fed into a feature pyramid network to obtain multi-scale features, namely $P_2$, $P_3$, $P_4$, $P_5$.
An extra $P_6$ is included for instance segmentation by down-sampling $P_5$.

\vspace{-10pt}
\paragraph{Instance segmentation branch.}
Instance segmentation branch enriches the features from $\{P_i | i = 2, \dots,6\}$ by ICM and performs instance segmentation at each level. For different levels, the weights of ICM are shared.
Our ICM is a compatible module that can easily be plugged into various one-stage instance segmentation methods.
In our work, we transplant the instance branch from pioneering SOLOv2~\cite{wang2020solov2} to perform instance segmentation. 

\vspace{-10pt}
\paragraph{Semantic segmentation branch.}
Semantic segmentation branch takes $\{P_i | i = 2 \dots 5\}$, augments them by SCM and performs semantic segmentation. For different levels, the weights of SCM are shared.
Following UPSNet~\cite{xiong2019upsnet}, we design a subnet consisting of 4 stacked $3\times3$ convolutions. This subnet is applied to each pyramid level. After that, the output features are upsampled and concatenated to predict semantic segmentation.

\vspace{-10pt}
\paragraph{Training and inference.}
To train our model, there are three loss terms in total:
\begin{equation}
	L = \underbrace{L_{\text{mask}} + L_{\text{cate}}}_{\text{instance segmentation loss}} + \underbrace{\lambda L_{\text{sem}}}_{\text{semantic segmentation loss}},
\end{equation}
where $\lambda$ is used to control the balance between tasks.
$L_{\text{mask}}$ and $L_{\text{cate}}$ are dice and focal loss following~\cite{wang2020solo, wang2020solov2}. $L_{\text{sem}}$ is cross entropy loss for semantic segmentation.
Parameters of SCM are optimized by semantic segmentation loss, while those of ICM are guided by instance segmentation loss. During inference, points with category scores higher than a certain threshold are first picked.
Afterwards, a Matrix-NMS~\cite{wang2020solo} is applied to remove duplicate predictions.
Finally, a heuristic algorithm~\cite{kirillov2019panoptic1} fuses instance and semantic segmentation results to obtain the final panoptic output.

\subsection{Correlation Functions}
\label{sec:3.2}
Here we introduce how to obtain the correlations applied in SCM and ICM.
Both instance and semantic correlations are pixel-wise and exist between a given location $i$ and all other locations.
Previous practice of utilizing appearance similarity or coordinates to compute correlations brings two drawbacks: 1). It does not take the image as a whole and ignore the continuity of the correlations of a specific location with other locations. 
2). It overlooks the image content that can be too rigid for various scenes. 
Hence, we propose the correlation function mechanism to directly model the underlying correlations as scene-relevant and tunes them to fit task needs through implicit supervisions. 

\begin{figure}[tb]
	\centering
	\includegraphics[width=\linewidth]{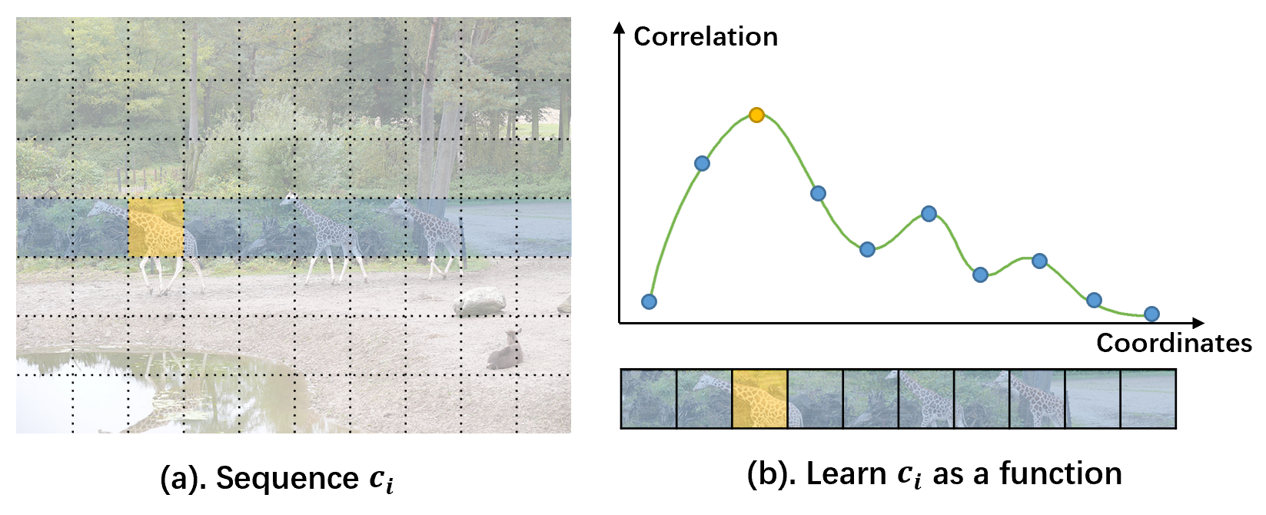}
	\vspace{-1.0em}
	\caption{
		Illustration of correlation function. 
		(a) demonstrates the concept of sequence $c_i$ in a real scenario, while (b) gives out a case of a possible sequence $c_i$ under this scenario and shows that a function of coordinates can be used to approximate it.
	}
	\label{fig:orthogonal}
	\vspace{-1.0em}
\end{figure}

\vspace{-10pt}
\paragraph{Sequence decomposition.}
Given a 1-dimensional input of length $L$, we denote the correlation between location $i$ and $j$ as $cor(i, j)$. Our target is to learn $cor(i, j)$ for arbitrary $i$ and $j$ with merely implicit task supervisions. This makes directly learning $cor(i, j)$ infeasible since optimizing each $cor(i, j)$ independently without explicit losses is extremely difficult. To alleviate this, we hope to organize relevant correlations together and learn them as a whole to capture their internal distribution as shown in Fig.~\ref{fig:orthogonal}. Specifically, we view the correlations between a given location $i$ and all other locations as a whole, and denote it by $C_i$:
\begin{equation}
	C_i = [cor(i, 0), ..., cor(i, L-1)].
\end{equation}
$C_i$ has the useful ``continuity'' property since the correlations between $i$ and any two adjacent locations ought to be close as nearby locations have similar contents. In other words, $C_i$ can be approximated as the output of discrete sampling on a hidden continuous function of locations. By doing so, the dense $L\times L$ correlation learning issue is decomposed to the learning of $L$ sequences $\{C_0, C_1, ..., C_{L-1}\}$, where each $C_i$ can be further learned as a whole by a approximation function.

\vspace{-10pt}
\paragraph{Sequence learning.}
We denote the approximation function of $C_i$ as $\mathcal{G}(\theta_i, j)$, where $\theta_i$ are parameters and $j$ are coordinates. The learning of variable-length $C_i$ is now converted to the learning of a fixed number of parameters $\theta_i$. Due to the fact that $C_i$ can be of arbitrary forms related to the specific scene content and task requirements, $\mathcal{G}$ must be powerful enough to cover any possible cases. 
Hence, we adopt DFT~(Discrete Fourier Transform), which is known for its capability to fit any discrete sequences. To meet its prerequisite, we first extend $C_i$ to a form a periodic signal $T_i$ of frequency $\omega=2\pi / L$ by having:
\begin{equation}
	T_i[j] = C_i[j \operatorname{mod} L].
\end{equation}

This extension leads to an issue that the beneficial ``continuity'' breaks between $T_i[L-1]$ and $T_i[L]$, where these two locations, \ie the start and end of the signal, tend to have different correlations with location i. To avoid this, we append $C_i$ with its mirrored signal to form $C^\prime_i$ before extending it to periodic signal $T_i^\prime(t)$:
\begin{equation}
	C_i^\prime [j]= \left\{
	\begin{array}{cc}
		C_i[j]       & {0 \le j < L}\\
		C_i[2L-j-1]     & {L \le j < 2L}.\\
	\end{array} \right.
\end{equation}

After obtaining the periodic discrete signal $T^\prime_i$ of frequency $\omega^\prime = \omega / 2$, DFT approximates it as an addition of $N$ base sinodual functions parameterized by amplitudes $A_n^i$ and phrases $\psi_n^i$ as follows:
\begin{align}
	\label{eqa:fourier}
	T_i^\prime(j) &\approx A_{0}^i +\sum_{n=1}^{N} A_{n}^i \sin \left(n \omega^\prime j + \psi_{n}^i \right) \stackrel{\text{def}}{=} \mathcal{G}(\theta_i, j),
\end{align}
where $\theta_i=\{A_0^i\} \cup \{A_n^i, \psi_n^i | n = 1, ..., N\}$. 
To utilize the knowledge that the concrete form of this function~(\ie sequence) varies with scenes, we choose to predict $\theta_i$ from signal contents. For $L$ sequences, there are $L\times (2N+1)$ parameters in total, where N is the level of expansion in Eq.~\ref{eqa:fourier}. This is implemented through conventional convolutions, where feature of each location $i$ is responsible to predict $\theta_i$ to generate the correlation sequence $C_i$. Afterwards, $C_i$ will be used along with signal content to fulfill tasks so that $\theta_i$ can be optimized under implicit supervisions.

\vspace{-10pt}
\paragraph{2-dimensional extensions.}
Extending the 1-dimensional signal of $L$ to 2-dimensional~(2d) inputs of $H\times W$ is non-trivial since the assumption of continuity breaks when row or column changes. Nevertheless, for locations belonging to the same row or column, such an assumption still exists. Hence, we choose to approximate the function of 2-dimensional locations as the product of two independent axial 1-dimensional functions:
\begin{equation}
	\label{eqa:2dcor}
	\begin{aligned}
		cor_{2d}(u, v) = &cor_{hor} (u, v) \cdot cor_{ver}(u, v) \\
		= &\mathcal{G}(v_x, \theta_u^{hor}) \cdot \mathcal{G}(v_y, \theta_u^{ver}),
	\end{aligned}
\end{equation}
where $u$, $v$ are two 2d locations, $v_x$ and $v_y$ are the coordinates of location $v$ along horizontal and vertical axis.
And $\theta_u = [\theta_u^{hor}, \theta_u^{ver}]$ are the parameters of two axial 1-dimensional functions. The learning of $HW \times HW$ correlations are now converted to $H \times W \times 2$ sets of parameters, where each contains $2N+1$ variables. Similarly, the total $H \times W \times 2(N+1)$ is predicted by applying convolutions upon input features and receives supervision from implicit tasks losses.

\vspace{-10pt}
\paragraph{Discussions with the decoupled SOLO~\cite{wang2020solo}.}
We decompose dense correlation predictions to two orthogonal function learning, while decoupled SOLO predicts two dense $1d$ tensors. However, the underlying motivations are distinct. Decoupled SOLO decomposes dense prediction for efficiency, while we are to make use of ``spatial locality'', i.e, we want adjacent locations of similar positional correlation. If we learn a single function to predict the $2d$ correlation, the end of row $i$ will get similar correlations as the start of row $i+1$ since they are considered ``adjacant'' in the flatten $1d$ view. ~(No matter how you flats, the ``false adjacant'' issue will still exist). One would not want that since the two ends of one image tend to be irrelevant.

\begin{figure}[tb]
\centering
\includegraphics[width=1.0\linewidth]{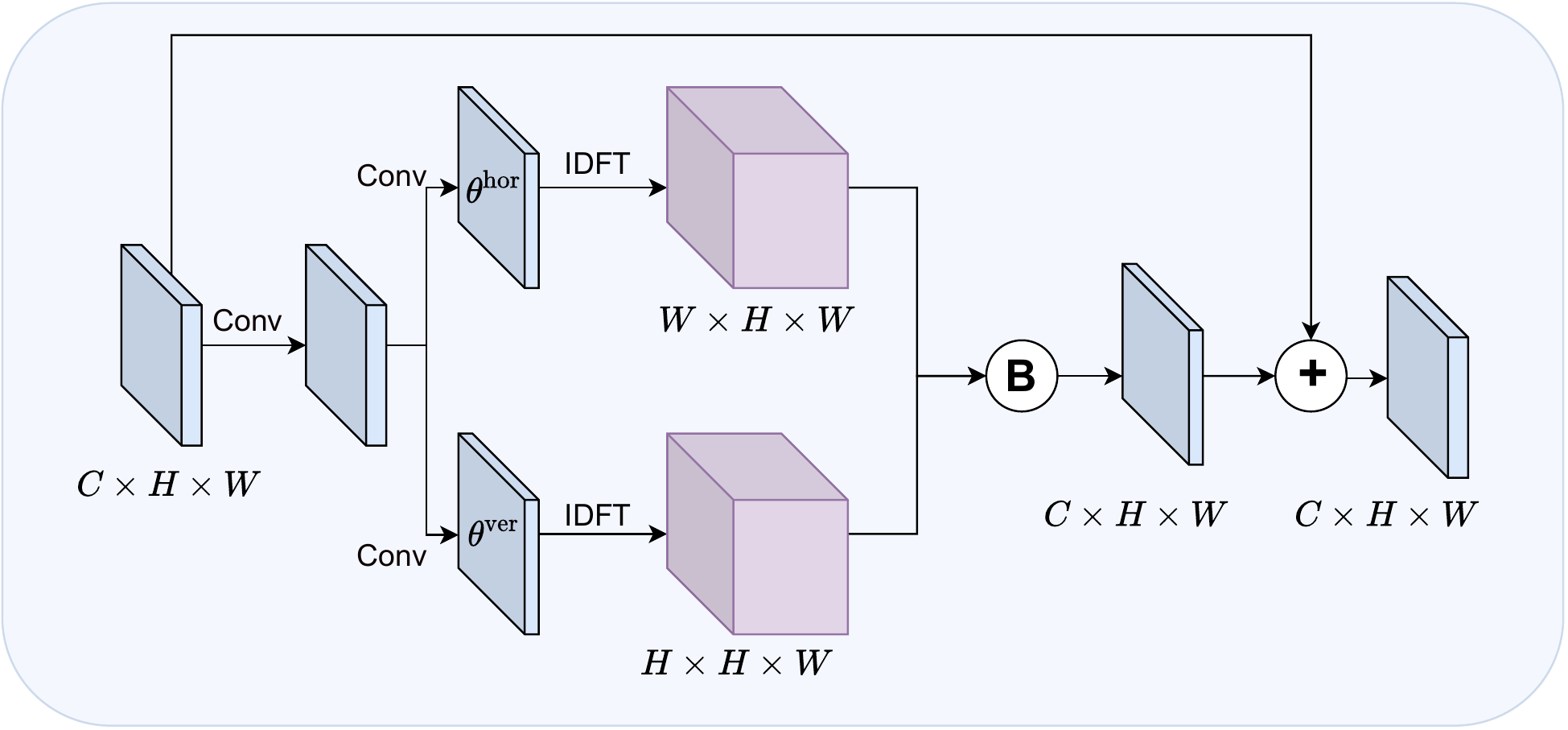}
\caption{
Structure of semantic correlation module. B represents the broadcast process in Eq.~\ref{eqa:axial}, and \textcircled{+} indicates element-wise addition.
}
\label{fig:sempos}
\vspace{-1.0em}
\end{figure}

\subsection{Application of Correlations}
\label{sec:3.3}
With the proposed correlation functions, we are able to model correlations directly from scene inputs. In this section, we introduce two modules, namely Semantic Correlation Module~(SCM) and Instance Correlation Module~(ICM), to make use of such correlation respectively in semantic segmentation and instance segmentation, two subtasks of panoptic segmentation.

\vspace{-10pt}
\paragraph{Semantic correlation module.}
To learn semantic-level discriminative features, the design of semantic correlation module~(SCM) follows the spirit of how attention mechanism~\cite{wang2020axial, wang2018location} are applied in semantic segmentation~\cite{vaswani2017attention}. Recall that in the original attention mechanism, one defines the correlation between one location $u$ and location $v$ as:
\begin{equation}
	\begin{aligned}
		w_{u, v} = f_u^T QK^T f_v = f_u^T M f_v,
	\end{aligned}
\end{equation}
where $M=Q K^T$,$Q$ and $K$ are projection matrices, $f_u$ and $f_v$ are visual features at location $u$ and $v$. 
Naturally, our SCM replaces $w_{u, v}$ with $cor_{\operatorname{2d}}(u, v)$, which can model scene-related, task-specific semantic-level correlation directly from scenes. It not only saves the trouble for time-consuming attention operations, but also enables the learning of complicated patterns. As shown in Fig.~\ref{fig:sempos}, our SCM applies a single $3\times 3$ convolution on the input features and predicts $\theta_u = [\theta_u^{hor}, \theta_u^{ver}]$ densely by two $1\times 1$ convolutions. Afterwards, $cor_{2d}(u, v)$ is calculated following Eq.~\ref{eqa:fourier} and Eq.~\ref{eqa:2dcor}.
As for feature propagation, we provide two modes, namely global aggregation and axial aggregation. The former one aggregates information from all locations as:
\begin{equation}
\begin{aligned}
	f_u^S = &\sum_{v} \operatorname{softmax}_v(cor_{2d} (u, v))  f_v 
\end{aligned}
\end{equation}
The latter one gathers information from axial locations as:
\begin{equation}
\label{eqa:axial}
\begin{aligned}
		f_u^S = &\sum_{v \in \{p| p_x = u_x\}} \operatorname{softmax}_v(cor_{hor} (u, v))  f_v \\
+& \sum_{v \in \{p| p_y = u_y\}} \operatorname{softmax}_v(cor_{ver} (u, v))  f_v.
\end{aligned}
\end{equation}
The enhanced feature $f_u^S$ is used to perform semantic segmentation while $\theta_u$ can be optimized through the supervision from semantic segmentation.
Taking efficiency into consideration, Axial aggregation
mode is chosen for SCM.

\begin{figure}[tb]
	\centering
	\includegraphics[width=1.0\linewidth]{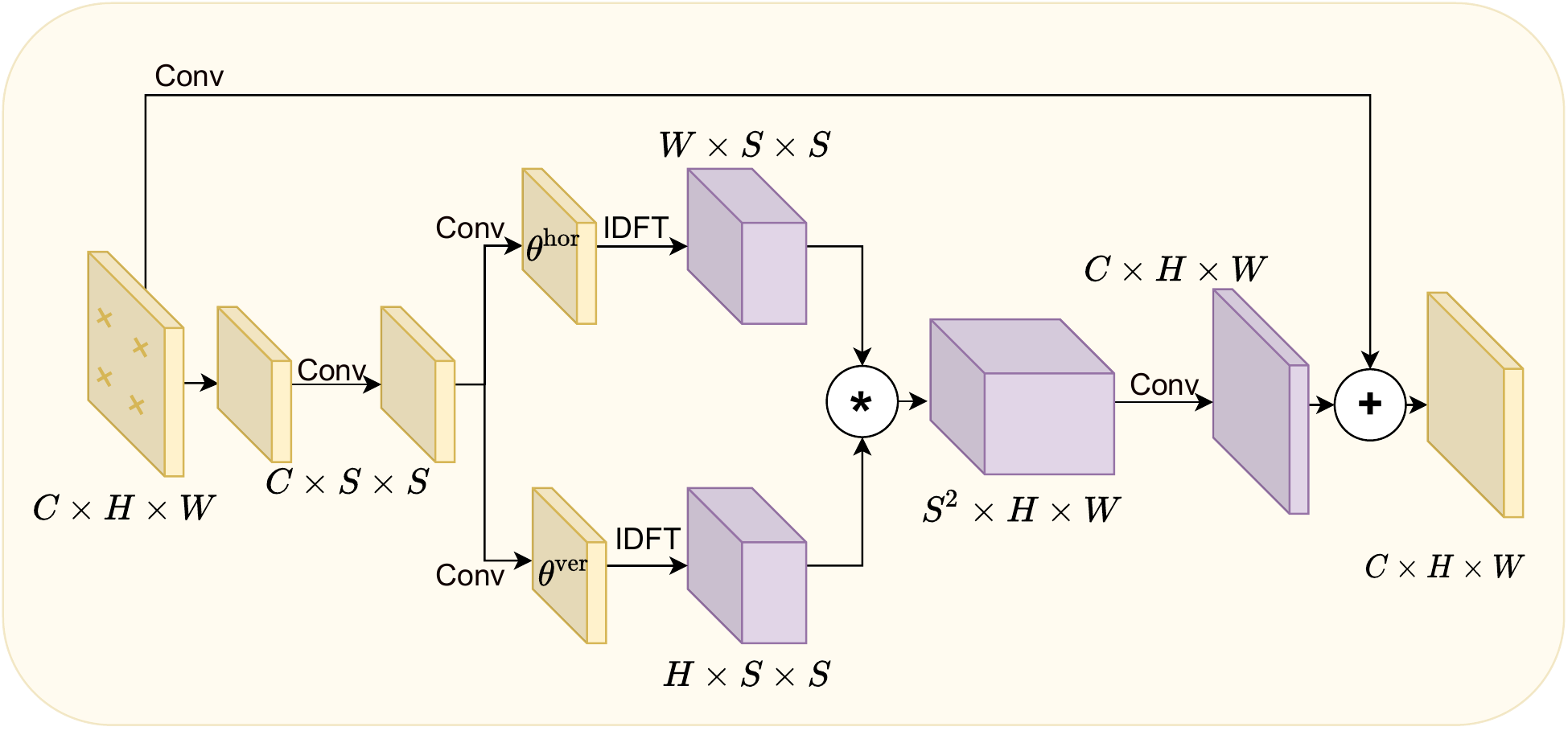}
	\caption{
		Structure of instance correlation module. $S^2$ reference points are uniformly sampled.
	}
	\label{fig:inspos}
	\vspace{-1.0em}
\end{figure}

\vspace{-10pt}
\paragraph{Instance correlation module.}
Directly adopting the above semantic correlation mechanism for instance segmentation is not reasonable, because the aggregated features for different instances belonging to the same category could still be very similar.
Here, we hope to achieve instance discrimination which means the modification for different instances should be different.
We find that if two locations are belonging to the same instance, then their correlations with other locations would be similar. 
Therefore, the correlations themselves could be viewed as discriminative features to make different instances distinguishable.
This can help instance segmentation to tell instances apart.
To be more specific, we concatenate the correlations of a given location and combine them with the visual features.
Our ICM combines concatenated correlations with semantic features using the same addition operations. We sample $C$ fixed locations as references, denoted as $P$, in the input and calculate the correlation of each location $u$ with respect to these $C$ reference locations as $c_u$:
\begin{equation}
	c_u = [cor_{2d}(u, P_1), ..., cor_{2d}(u,P_C)] ^ T. 
\end{equation}
Afterwards, it is directly added to visual features as:
\begin{equation}
	f_u^I = W_f^T f_u + c_u
\end{equation}

In practice, instead of sampling all points, we choose $S\times S$ points instead for efficiency.
To match $S^2$ with $C$, a further $1\times 1$ convolution is applied. The whole process of instance correlation module is illustrated in Fig.~\ref{fig:inspos}. The enhanced feature $F^I$ is later used to perform instance segmentation.

\begin{table*}[tb]
	\centering
	\footnotesize
	\begin{tabular}{ l l c c c c c c c c c c}
		\toprule
		method & Backbone & $\text{PQ}$ & $\text{SQ}$ & $\text{RQ}$ & $\text{PQ}^\text{Th}$ & $\text{SQ}^\text{Th}$ & $\text{RQ}^\text{Th}$ & $\text{PQ}^\text{St}$ & $\text{SQ}^\text{St}$ &  $\text{RQ}^\text{St}$ & FPS \\
		\hline
		\multicolumn{12}{c}{\textit{two-stage}} \\
		\hline
		Panoptic-FPN~\cite{kirillov2019panoptic2} & Res50-FPN  & 39.0  & - & - & 45.9 & - &  - & 28.7 & - & -  & - \\
		UPSNet~\cite{xiong2019upsnet} & Res50-FPN &  42.5  & 78.0 & 52.5 & 48.6 & 79.4 & 59.6 & 33.4 & 75.9 & 41.7  & 9.1 \\
		BANet~\cite{chen2020banet} & Res50-FPN &  43.0  & 79.0 & 52.8 & 50.5 & 81.1 & 61.5 & 31.8 & 75.9 & 39.4 &  - \\
		BRGNet~\cite{wu2020bidirectional} & Res50-FPN  & 43.2  & - & - & 49.8 & - & -& 33.4 & - & -&  - \\
		Unifying~\cite{li2020unifying} & Res50-FPN &  43.4   & {79.6} & 53.0 & 48.6 & - & -& 35.5 & - & - &  - \\
		CIAE~\cite{gao2020learning} & Res50-FPN   &  40.2 & - & - & 45.3 & - & - & 32.3 & - & - & 12.5 \\
		SOGNet~\cite{yang2019sognet} & Res50-FPN &  {43.7}  & 78.7 & {53.5} & {50.6} & - & - & 33.2 & - & - & - \\
		\hline
		\multicolumn{12}{c}{\textit{one-stage}} \\

		\hline
		DeeperLab~\cite{yang2019deeperlab} & Xception-71  & 33.8   & -   & - & - & - & -& - & - & - & 10.6 \\
		RealTimePan~\cite{hou2020real} & Res50-FPN &  37.1  & - & - & 41.0  & - & - & 31.3  & - & -& \textbf{15.9} \\
		PCV~\cite{wang2020pixel} & Res50-FPN &  37.5 & 77.7 & 47.2 & 40.0  & - & - & 33.7  & - & - & 5.7 \\
		Panoptic-Deeplab~\cite{chen2019pdeeplab} & Xception-71  & 39.7  & - & - & 43.9   & - & - & 33.2  & - & - & 7.6 \\
		SOLOv2~\cite{wang2020solov2} & Res50-FPN &  42.1 & - & -  & 49.6 & - & - & 30.7 & - & - & -\\
		LPSNet~\cite{hong2021lpsnet} & Res50-FPN &  42.4 & 79.2 & 52.7  & 48.0 & 35.8 & - & - & - & - & 6.8\\
		Panoptic-FCN~\cite{li2021panopticfcn} &  Res50-FPN &   43.6 & 80.6 & 52.6& 49.3 & 82.6 & 58.9 & 35.0 & 77.6 & 42.9 & 12.5 \\
		$\text{Panoptic FCN}^\star$~\cite{li2021panopticfcn} &  Res50-FPN &   44.3 & 80.7 & 53.0& 50.0 & \textbf{83.4} & 59.3 & \textbf{35.6} & 76.7 & 43.5 & 9.2 \\
		\hline
		$\text{Ours}$ & Res50-FPN &  44.5 & 79.8 & 54.0 & \textbf{52.1}& 82.8 & \textbf{62.1} & 33.1 & 75.2 & 41.7 & 13.3 \\
		$\text{Ours}^\star$ & Res50-FPN &  \textbf{45.1} & \textbf{81.0} & \textbf{54.4} & 51.8 & 82.8 & 61.7 & 35.1 & \textbf{78.2} & \textbf{43.5} & 10.5 \\
		\bottomrule
	\end{tabular}
	\caption{Results on COCO \emph{val} set.}
	\label{tab:sota}
\vspace{-0.5em}
\end{table*}

\section{Experiments}
In this section, we evaluate our methods on challenging benchmarks COCO~\cite{lin2014microsoft} and ADE20k~\cite{zhou2017ade}. In addition, we perform extensive ablation studies to verify the effectiveness of different components.

\vspace{-10pt}
\paragraph{Datasets.}
MS COCO is a large-scale panoptic segmentation dataset with 118k training images, 5k validation images, and 20k test images. Annotations of 80 thing categories and 53 stuff categories for both instance segmentation and semantic segmentation are provided. We train our model on the \textit{train} set without extra data and report results on both \textit{val} and \textit{test-dev} sets. ADE20K is a densely annotated dataset for panoptic segmentation containing 20k images for training, 2k images for validation and 3k images for test. It contains 150 classes in total, including 100 things and 50 stuff classes. We train our model on its \textit{train} set without extra data and report results on its \textit{val} set.
We utilize PQ~\cite{kirillov2019panoptic1}, which is averaged over categories, to evaluate the panoptic segmentation performance. PQ is the combination of recognition quality~(RQ) and segmentation quality~(SQ).
$\text{PQ}^{\text{Th}}$ averaged over thing categories and $\text{PQ}^\text{St}$ averaged over stuff categories are also reported to reflect the improvement on instance and semantic segmentation.

\begin{figure}[tb]
	\centering
	\includegraphics[width=\linewidth]{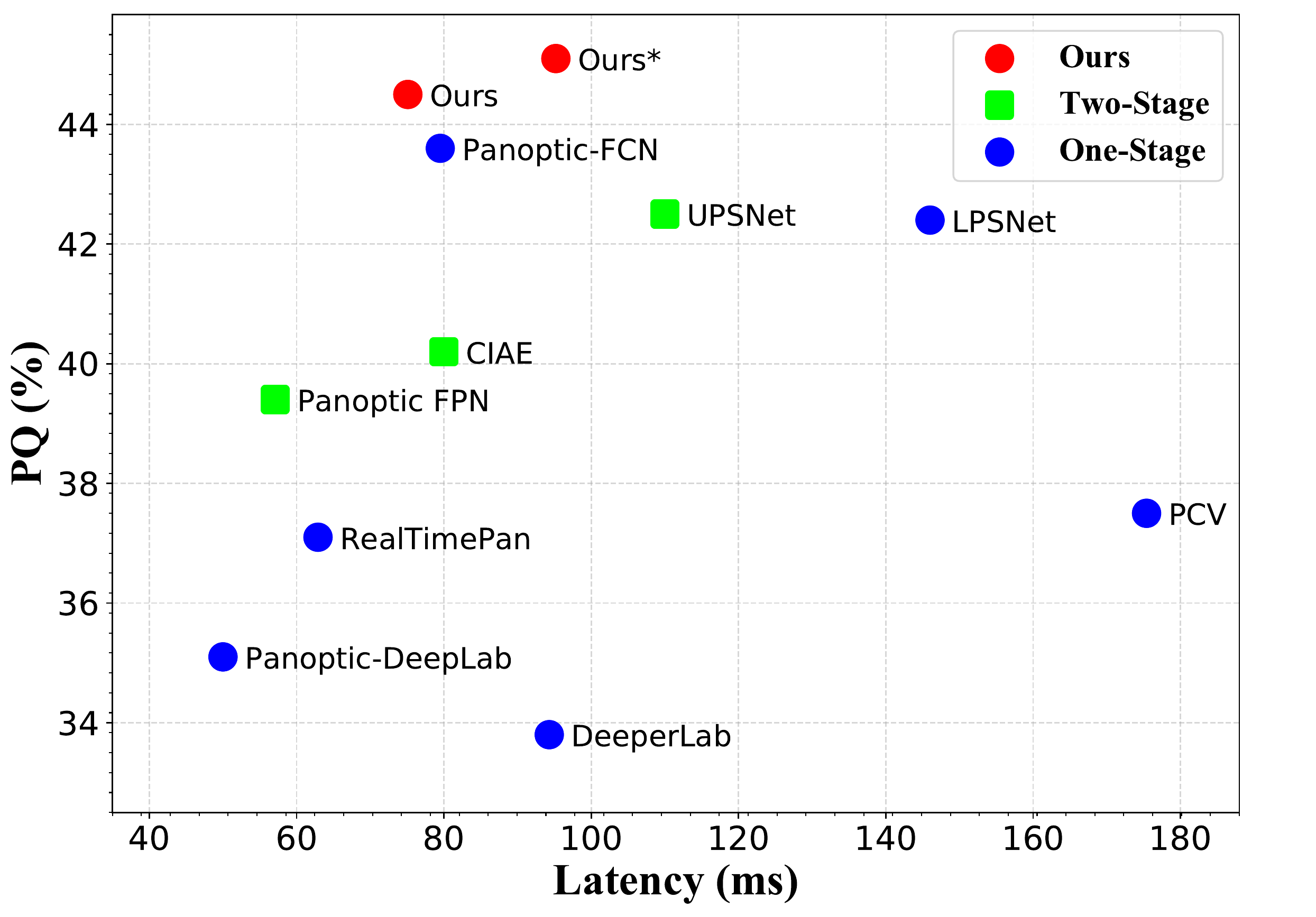}
	\vspace{-1.0em}
	\caption{
		PQ and latency on COCO \textit{val} set. Our method achieves state-of-the-art performance while keeping a competitive inference speed.
	}
	\label{fig:latency}
	\vspace{-1.0em}
\end{figure}

\vspace{-10pt}
\paragraph{Implementation details.}
We choose Res50-FPN and Dcn101-FPN as our backbone for val and test-dev respectively. We set Fourier decomposition level $N$ to 5 and reference spatial size $S$ to 16 to well balance performance and efficiency. $\lambda$ is set to 0.5. For fair comparison on MS COCO, we report performances obtained from 36 epochs training schedule following~\cite{li2021panopticfcn}. We use the SGD optimization algorithm with momentum of 0.9 and weight decay of 1e-4. The batch-size is set to 16. For data augmentation, random crop and horizontal flip are used. As for inference, the score thresholds before and after NMS are set to 0.1 and 0.3. Stuff regions whose areas are below 4096 are removed~\cite{kirillov2019panoptic1}. For ADE20k, we merely adjust the training schedule. Following the setting of BGRNet~\cite{wu2020bidirectional}, we train for 24 epochs with learning rate divided by 10 at the 18 and 22 epochs.

\begin{table}[tb]
	\centering
	\footnotesize
	\begin{tabular}{ l  l  c c c }
		\toprule
		Method & Backbone & $\text{PQ}$ & $\text{PQ}^\text{Th}$ & $\text{PQ}^\text{St}$ \\
		\hline
		\multicolumn{5}{c}{\textit{two-stage}} \\
		\hline
		Panoptic-FPN & Res101-FPN & 40.9 & 48.3 & 29.7\\
		AUNet & ResX152-FPN & 46.5 & \textbf{55.9} & 32.5 \\
		UPSNet & Dcn101-FPN & 46.6 & 53.2 & 36.7 \\
		Unifying & Dcn101-FPN & 47.2 & 53.5 & 37.7 \\
		BANet & Dcn101-FPN & 47.3 & 54.9 & 35.9 \\
		SOGNet & Dcn101-FPN & 47.8 & - & - \\
		\hline
		\multicolumn{5}{c}{\textit{one-stage}} \\
		\hline
		DeeperLab  & Xception-71 & 34.3 & 37.5 & 29.6 \\
		SSAP  & Res101-FPN & 36.9 & 40.1 & 32.0 \\
		Panoptic-Deeplab & Xception-71 & 39.6 & -& - \\
		Axial-Deeplab & Axial-ResNet-L & 43.6 & 48.9 & {35.6} \\
		Panoptic-FCN$^\star$ & Dcn101-FPN & 47.5 & 53.7 & \textbf{38.2} \\
		\hline
		$\text{Ours}^\star$ & Dcn101-FPN & \textbf{48.1} & 55.6 & 36.7 \\
		\bottomrule
	\end{tabular}
	\caption{Results on COCO \emph{test-dev} set.}
	\label{tab:sota_test}
	\vspace{-1.0em}
\end{table}

\subsection{Comparison with State-of-the-Art Methods}
We compare our method with state-of-the-art methods on COCO \textit{val} and \textit{test-dev} on the GPU device NVIDIA V100.
Quantitative results on \textit{val} are shown in Tab.~\ref{tab:sota} and Fig.~\ref{fig:latency}. We provide results without/with deformable convolutions, denoted as Ours and $\text{Ours}^\star$ respectively. As one can tell in Tab.~\ref{tab:sota}, $\text{Ours}^\star$ based on Res50-FPN outperforms all the comparing single-stage methods in both accuracy and efficiency on COCO \textit{val}. When compared to Panoptic-FCN, our method is 0.9\% PQ higher with a higher running speed~(+0.8FPS). With deformable convolutions, $\text{ours}^\star$ exceeds two-single stage methods by 1.4\% PQ and is still faster.
In Tab.~\ref{tab:sota_test}, we provide results on COCO \textit{test-dev} subset. $\text{Ours}^\star$ based on Dcn101-FPN outperforms all top-listed approaches and even exceeds the best two-stage SOGNet~\cite{yang2019sognet} by 0.3\% PQ, showing the great potential of our method combined with a powerful backbone.

In Tab.~\ref{tab:sota_ade}, we compare our method with state-of-the-art methods on ADE20k \textit{val}-set. Being the first single-stage method to provide results an ADE20k, our method outperforms previous two-stage methods by 0.8\% higher PQ. It shows a stronger performance for both foreground and background regions, showing that our method is able to learn correlations effectively for different datasets.

\subsection{Ablation Study}
We perform ablation studies on COCO \textit{val} with our model based on Res50-FPN. For efficiency, we train the model with a shorter training schedule (12 epochs) following~\cite{li2021panopticfcn} and do not employ Deformable Convolution.
At first, we study the effectiveness of our correlation modules by adding them one-by-one to the baseline which directly attaches instance and semantic heads to the backbone features. Next, we study the importance of adapted correlation information by comparing them to existing CoordConv~\cite{liu2018intriguing} and attention mechanism~\cite{wang2018non} on separate segmentation tasks.
%
Finally, we give ablations on the choice of hyper-parameters and computational cost.

\begin{table}[t!]
	\centering
	\footnotesize
	\begin{tabular}{ l  l  c c c }
		\toprule
		Method & Backbone & $\text{PQ}$ & $\text{PQ}^\text{Th}$ & $\text{PQ}^\text{St}$ \\
		\hline
		\multicolumn{5}{c}{\textit{two-stage}} \\
		\hline
		Panoptic-FPN & Res50-FPN & 29.3 & 32.5 & 22.9 \\
		$\text{Panoptic-FPN}^\dag$ & Res50-FPN & 30.1 &33.1 &24.0 \\
		BGRNet & Res50-FPN & 31.8 & 34.1 & 27.3 \\
		\hline
		\multicolumn{5}{c}{\textit{one-stage}} \\
		\hline
		$\text{Ours}^\star$ & Res50-FPN & \textbf{32.6} & \textbf{35.0} & \textbf{27.9} \\
		\bottomrule
	\end{tabular}
	\caption{Results on ADE20k \emph{val} set.}
	\label{tab:sota_ade}
	\vspace{-1.0em}
\end{table}

\begin{table}[tb]
	
	\centering
	\footnotesize
	\begin{tabular}{c c  c c c  c}
		\toprule
		SCM & ICM & $\text{PQ}$ & $\text{SQ}$ & $\text{RQ}$ &  Inf~(ms)  \\
		\hline
		& & 39.3 & 77.6 & 48.3& \textbf{84.3}\\
		$\checkmark$& & 39.9& 77.8& 49.1 & 88.2\\
		& $\checkmark$& 39.6 & 77.1 & 48.7  & 90.0\\
		$\checkmark$ & $\checkmark$ & \textbf{40.5}& \textbf{78.6}& \textbf{49.7}& 93.4 \\
		\bottomrule
	\end{tabular}
	\caption{Ablation Study on COCO \emph{val} set.}
	\label{tab:ablmodules}
\vspace{-1em}
\end{table}

\vspace{-10pt}
\paragraph{Semantic correlation module.}
To study the effect of semantic correlation module~(SCM), we run experiments with SCM alone.
As shown in the second row of Tab.~\ref{tab:ablmodules}, applying SCM alone leads to a 0.6\% gain in PQ. This mainly comes from the improved $\text{PQ}^\text{St}$~(+2.7\%), showing that SCM is able to learn beneficial semantic correlation under the implicit supervision from semantic branch.

\vspace{-10pt}
\paragraph{Instance correlation module.}
To study the effect of instance correlation module~(ICM), we run experiments with ICM alone and with both ICM and SCM. As shown in the third row of Tab.~\ref{tab:ablmodules}, applying ICM alone leads to a 0.3\% gain in terms of PQ. Such an increment can be attributed to the $0.5\%$ higher $\text{PQ}^\text{Th}$. This demonstrates that ICM effectively models the correlation information suitable for instance segmentation.
Since ICM needs to predict instance-level correlations from the backbone features and thus incorporating ICM alone into the one-stage framework may encourage the backbone features to become more instance-level discriminative and affect the semantic segmentation branch.
More importantly, by jointly applying SCM and ICM, our performance is further increased by 0.6\%. This reveals that our correlation modules are able to work together to provide task-specific scene-adaptive discriminative features to alleviate the intrinsic conflicts between two segmentation tasks, thus leading to better performance.

\begin{figure*}[tb]
	\centering
	\includegraphics[width=0.95\linewidth]{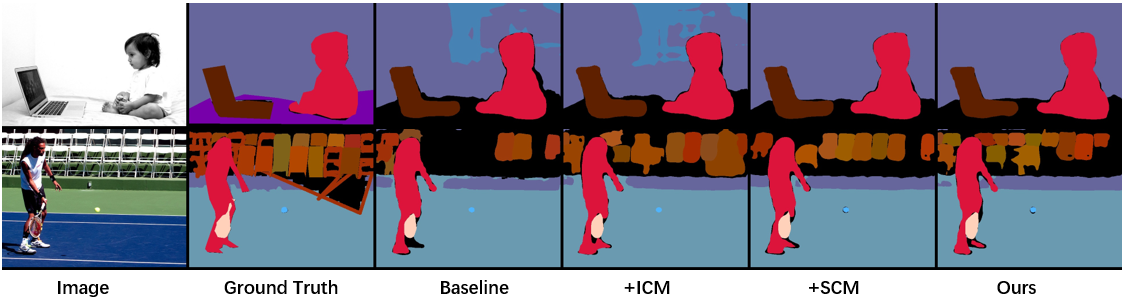}
	\vspace{-0.5em}
	\caption{
		Qualitative results of different correlation modules on COCO.
	}
	\label{fig:vis_output}
\vspace{-0.5em}
\end{figure*}


\begin{figure}[tb]
	\centering
	\vspace{-0.5em}
	\includegraphics[width=0.9\linewidth]{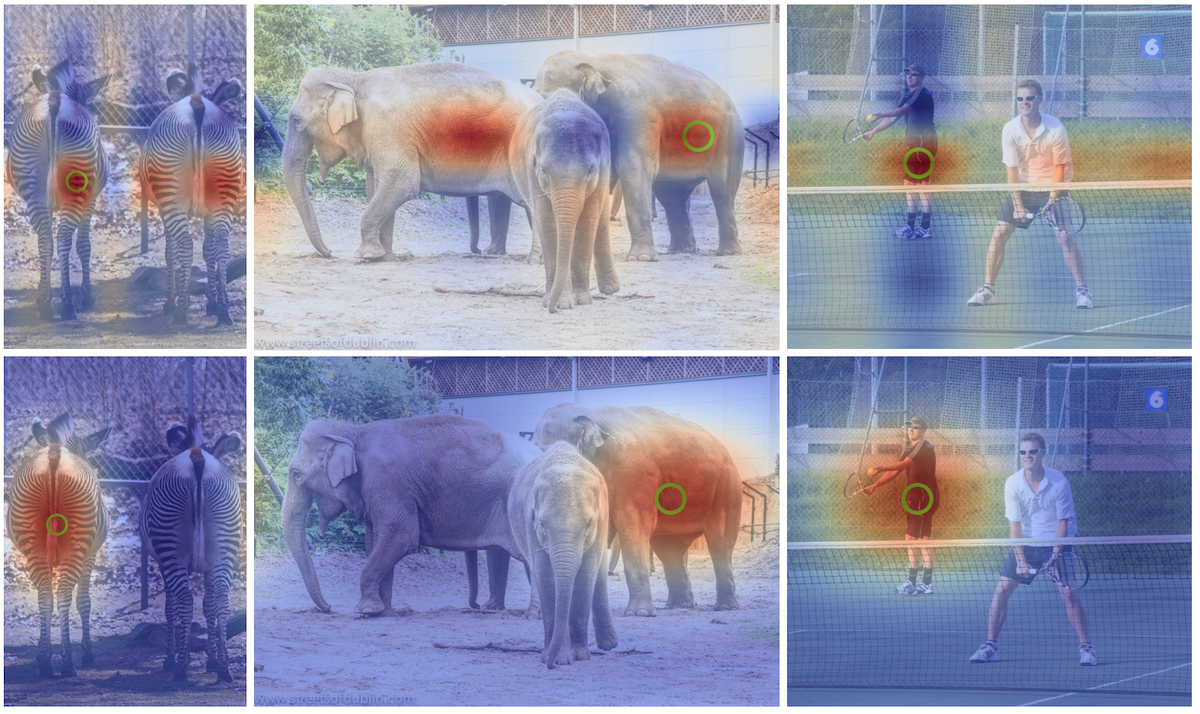} 
	\caption{
		Visualizations of the implicitly learned semantic (first row) and instance (second row) correlations. Each correlation map is obtained by calculating the correlation between each location and a fixed location~(labeled as a green point) with a warmer color for a stronger correlation.
	}
	\label{fig:supp_cor}
\end{figure}

\begin{table}[t]
	\centering
	\footnotesize
	\begin{tabular}{c | c |c | c}
		\toprule
		methods & - & att & SCM\\
		\hline
		mIoU  & 47.6 & 50.1 & \textbf{50.7} \\
		\bottomrule
	\end{tabular}
	\caption{Ablation on SCM on COCO \emph{val} set.}
	\label{tab:att}
\end{table}

\begin{table}[t]
	\centering
	\footnotesize
	\begin{tabular}{c | c |c | c | c}
		\toprule
		methods & - & coords & sinoduals & ICM\\
		\hline
		mIoU  & 29.5 & 32.9 & 32.5 & \textbf{33.5} \\
		\bottomrule
	\end{tabular}
	\caption{Ablation on ICM on COCO \emph{val} set.}
	\label{tab:ablsem}
\end{table}

\begin{table}[t]
	\centering
	\footnotesize
	\begin{tabular}{c c c c| c c c}
		\toprule
		& \multicolumn{3}{c|}{Semantic}& \multicolumn{3}{c}{Instance(N=5)}\\
		\hline
		& N=3 & N=5 & N=7 & S=8 & S=16 & S=32  \\
		\hline
		mIoU & 50.3 & \textbf{50.7} & 50.5 & - & - & - \\
		\hline
		mAP & - & - & -& 33.3 & 33.5 & \textbf{33.6} \\
		\bottomrule
	\end{tabular}
	\caption{Ablation on hyper-parameters on COCO \emph{val} set.}
	\label{tab:ablK}
\vspace{-1em}
\end{table}

\vspace{-10pt}
\paragraph{Task-driven and Scene-related correlation.}
To show the importance of task-driven and scene-related correlation information, we compare SCM and ICM with CoordConv~\cite{liu2018intriguing} and attention mechanism~\cite{wang2018non} on instance and semantic segmentations separately.
For semantic segmentation, the baseline is a backbone network with a segmentation head of 3 stacked $3\times3$ convs. And we incorporate self-attention~\cite{wang2018non} and SCM modules into the segmentation head in baseline denoted as ``att'' and ``SCM'', respectively. 
As shown in Table~\ref{tab:att}, our method SCM outperforms ``att'' with a noticeable gain.
For instance segmentation, the self-attention module may enhance the feature similarity for those locations belonging to the same semantic category and thus hinder instance-level discrimination. Therefore, we compare with coordinates~\cite{liu2018intriguing} and sinodual embeddings~\cite{carion2020end} based methods instead of self-attention methods.
As shown in Tab.~\ref{tab:ablsem}, our ICM obtains $33.5\%$ mAP, 0.6\% higher than using coordinates.

\vspace{-10pt}
\paragraph{Hyper-parameters.}
There are two hyper-parameters in our proposed modules, namely the Fourier decomposition level $N$ and spatial size of reference points $S$. We conduct ablation studies on these two hyper-parameters separately on semantic and instance segmentation. As shown in Tab.~\ref{tab:ablK}, the highest performance is achieved on semantic segmentation when $N$ is set to 5. The DFT function could not approximate the correlations well when the $N$ is too small. And a higher $N$ may lead to overfitting. For instance segmentation, $S=32$ is slightly better than $S=16$. Taking efficiency into account, we set $S$ to $16$ in all experiments.

\vspace{-10pt}
\paragraph{Ablation on computational cost.}
In this part, we perform an ablation study to demonstrate that the proposed method is more effective than using more extra computations or parameters.
The baseline is based on Res50-FPN with a standard SOLO instance segmentation head and a segmentation head of 3 stacked $3\times3$ convs. %
To show the performance gain of our method is not obtained by merely adding more computations or parameters, we make the segmentation head in baseline deeper to 7 convs, denoted as ``Deeper Head''.
As shown in Table~\ref{tab:deeper}, our method, with a similar runtime, outperforms ``Deeper Head'' with a noticeable gain w.r.t. \textbf{all metrics} and uses less parameters.

\begin{table}[t]
	\caption{Ablation on extra parameters on COCO \emph{val} set.}
	\label{tab:deeper}
	\vspace{-0.5em}
	\centering
	\footnotesize
	\begin{tabular}{c | c c  c | c | c}
		\toprule
		methods & $\text{PQ}$ & $\text{SQ}$ & $\text{RQ}$ & Inf~(ms) & Params~(M)\\
		\hline
		Baseline  & 39.3 & 77.6 & 48.3  &  \textbf{84.3}  \ & \textbf{37.55} \\
		Deeper Head  & 39.9 & 78.2 & 48.9  &  93.3  \ & 38.88(+1.33) \\
		Ours & \textbf{40.5} & \textbf{78.6} & \textbf{49.7}  & 93.4 \ &  38.15(+0.60) \\
		\bottomrule
	\end{tabular}
\end{table}
 
\subsection{Visualizations}
In this part, we provide some qualitative results to demonstrate the effectiveness of the proposed algorithm. We first demonstrate visual examples of the panoptic segmentation results obtained by our method in Fig.~\ref{fig:vis_output}. In the first row, misclassified regions are corrected due to SCM. In the second row, ICM helps to identify missing instances, resulting in a higher $\text{RQ}^\text{Th}$.
In addition, we give more visualizations of correlations learned implicitly by the proposed method. As one can tell in Fig.~\ref{fig:supp_cor}, our algorithm is able to capture task-specific and scene-related correlations for different segmentation tasks.
%

\section{Conclusion}
In this paper, we propose a novel approach for learning discriminative features via correlation functions for scene understanding tasks.
It explicitly models the correlations as scene-relevant functions of locations and optimizes them though implicit supervisions.
We further design two modules, namely semantic correlation module and instance correlation module to apply correlations in various segmentation tasks.
These modules are integrated into an one-stage pipeline for efficient panoptic segmentation.
The experimental results demonstrate that the proposed method is able to learn task-specific correlations flexibly, leading to state-of-the-art performances on COCO and ADE20k.

{\small
\bibliographystyle{ieee_fullname}
\bibliography{main}
}

\clearpage

\end{document}